\documentclass[10pt]{article}
\usepackage[utf8]{inputenc}
\usepackage[T1]{fontenc}
\usepackage{amsmath}
\usepackage{amsfonts}
\usepackage{amssymb}
\usepackage[version=4]{mhchem}
\usepackage{stmaryrd}
\usepackage{hyperref}
\hypersetup{colorlinks=true, linkcolor=blue, filecolor=magenta, urlcolor=cyan,}
\usepackage{graphicx}
\usepackage[export]{adjustbox}
\graphicspath{ {./images/} }
\usepackage{caption}

\title{JEL: A Novel Model Linking Knowledge Graph entities to News Mentions }

\author{Michael Kishelev\\
J.P.Morgan\\
Jersey City, NJ, US\\
michael.kishelev@jpmorgan.com\\
\and
Pranab Bhadani\\
J.P.Morgan\\
Jersey City, NJ, US\\
pranab.bhadani@jpmchase.com \\
\and
Vinay Chaudhri\\
J.P.Morgan\\
Palo Alto, CA, US\\
vinay.chaudhri@jpmchase.com \\
\and
Wanying Ding\\
J.P.Morgan\\
Palo Alto, CA, US\\
wanying.ding@jpmchase.com}

\date{}

\let\svthefootnote\thefootnote
\newcommand\blfootnotetext[1]{%
  \let\thefootnote\relax\footnote{#1}%
  \addtocounter{footnote}{-1}%
  \let\thefootnote\svthefootnote%
}

\let\svfootnotetext\footnotetext
\renewcommand\footnotetext[2][?]{%
  \if\relax#1\relax%
    \ifnum\value{footnote}=0\blfootnotetext{#2}\else\svfootnotetext{#2}\fi%
  \else%
    \if?#1\ifnum\value{footnote}=0\blfootnotetext{#2}\else\svfootnotetext{#2}\fi%
    \else\svfootnotetext[#1]{#2}\fi%
  \fi
}

\begin{document}
\maketitle
\captionsetup{singlelinecheck=false}

\begin{abstract}
We present JEL, a novel computationally efficient end-to-end multineural network based entity linking model, which beats current state-of-art model. Knowledge Graphs have emerged as a compelling abstraction for capturing critical relationships among the entities of interest and integrating data from multiple heterogeneous sources. A core problem in leveraging a knowledge graph is linking its entities to the mentions (e.g., people, company names) that are encountered in textual sources (e.g., news, blogs., etc) correctly, since there are thousands of entities to consider for each mention. This task of linking mentions and entities is referred as Entity Linking (EL). It is a fundamental task in natural language processing and is beneficial in various uses cases, such as building a New Analytics platform. News Analytics, in JPMorgan, is an essential task that benefits multiple groups across the firm. According to a survey conducted by the Innovation Digital team ${ }^{1}$, around 25 teams across the firm are actively looking for news analytics solutions, and more than $\$ 2$ million is being spent annually on external vendor costs. Entity linking is critical for bridging unstructured news text with knowledge graphs, enabling users access to vast amounts of curated data in a knowledge graph and dramatically facilitating their daily work. This paper proposes a novel end-toend neural entity linking model that goes beyond vanilla string matching and leverages surface and context information to achieve a more accurate entity linking results. The main contributions of this paper can be concluded as: (1) We created a novel entity linking model that outperforms about $15 \%$ compared with the current state-of-the-art model, BLINK, which Facebook published in 2020. (2) We built up an end-to-end pipeline to provide a one-stop service to multiple types of entity linking tasks. (3) We released an SDK inside JPMC, which is available to any team who needs the service.
\end{abstract}

\section*{ACM Reference Format:}
Wanying Ding, Pranab Bhadani, Vinay Chaudhri, and Michael Kishelev. 2022. JEL: A Novel Model Linking Knowledge Graph entities to News Mentions. In Proceedings of Knowledge Graphs at JPMorgan AI Summit 2022

\footnotetext{${ }^{1}$ As reported in the News Analytics Strategy Deck by the Digital Innovation team

AISummit, Nov 08-09, 2022, Brooklyn,NY\\
© 2022\\
ACM ISBN XXX-X-XXXX-XXXX-X/XX/XX...\$XX. 00\\
\href{https://doi.org/10.1145/nnnnnnn.nnnnnnn}{https://doi.org/10.1145/nnnnnnn.nnnnnnn}
}
(AISummit). ACM, New York, NY, USA, 5 pages. \href{https://doi.org/10.1145/}{https://doi.org/10.1145/} nnnnnnn.nnnnnnnn

\section*{1 INTRODUCTION}
News Analytics is a firm-wide product aggregating news content and delivering powerful analytics, helping users across JPMorgan discover relevant and curated news that matters to their businesses. Understanding the news is becoming increasingly crucial to run businesses successfully. Traditional news analytics requires users to have rich prior knowledge about the data and conduct extensive document exploration for specific tasks. By contrast, Knowledge Graphs provides more convenience for news analytics. Knowledge Graphs put data in context through linking and semantic metadata, providing a powerful framework for data integration, analytics, unification, and sharing.

Figure 1 shows an example using a knowledge graph for news analytics. As mentioned in one piece of news, Hertz filed for bankruptcy due to the pandemic, and many JPMC clients could feel stressed as they are suppliers to Hertz. Such stress can pass down into its supply chain and trigger financial difficulties for other clients. JPMC may face different levels of risks from suppliers with different orders in Hertz's supply chain. With identifying "Hertz" mentioned in financial news as "Hertz Global Holdings Inc" in the knowledge graph (distinguished from "Hertz Furniture System", "Hertz Enterprise LLC", etc.), we can accurately track down Hertz's supply chain, identify stressed suppliers with different revenue exposure, and measure our primary risk due to Hertz's bankruptcy. Once stressed clients with significant exposure are detected, alerts can be sent out to corresponding credit officers.

Example in Figure 1 clearly shows us how a knowledge graph can assist better news analytics by connecting entities with semantic links. However, one may notice that in order to leverage a knowledge graph for news analysis, one critical step is to bridge the mentions ("Hertz" in the news) with an entity ("Hertz Global Holdings Inc") in the knowledge graph, which is called Entity Linking. Imagine if "Hertz" was linked with incorrect entities, it would result in too many false alarms, resulting in wasted effort.

However, Entity Linking is challenging due to name variations and entity ambiguity. An entity may have multiple surface forms, such as its full name, partial name, aliases, and abbreviations. For instance, the named entity "Cornell University" has its abbreviation "Cornell," and the named entity of "New York City" has its nickname

\begin{figure}[h]
\begin{center}
  \includegraphics[width=\textwidth]{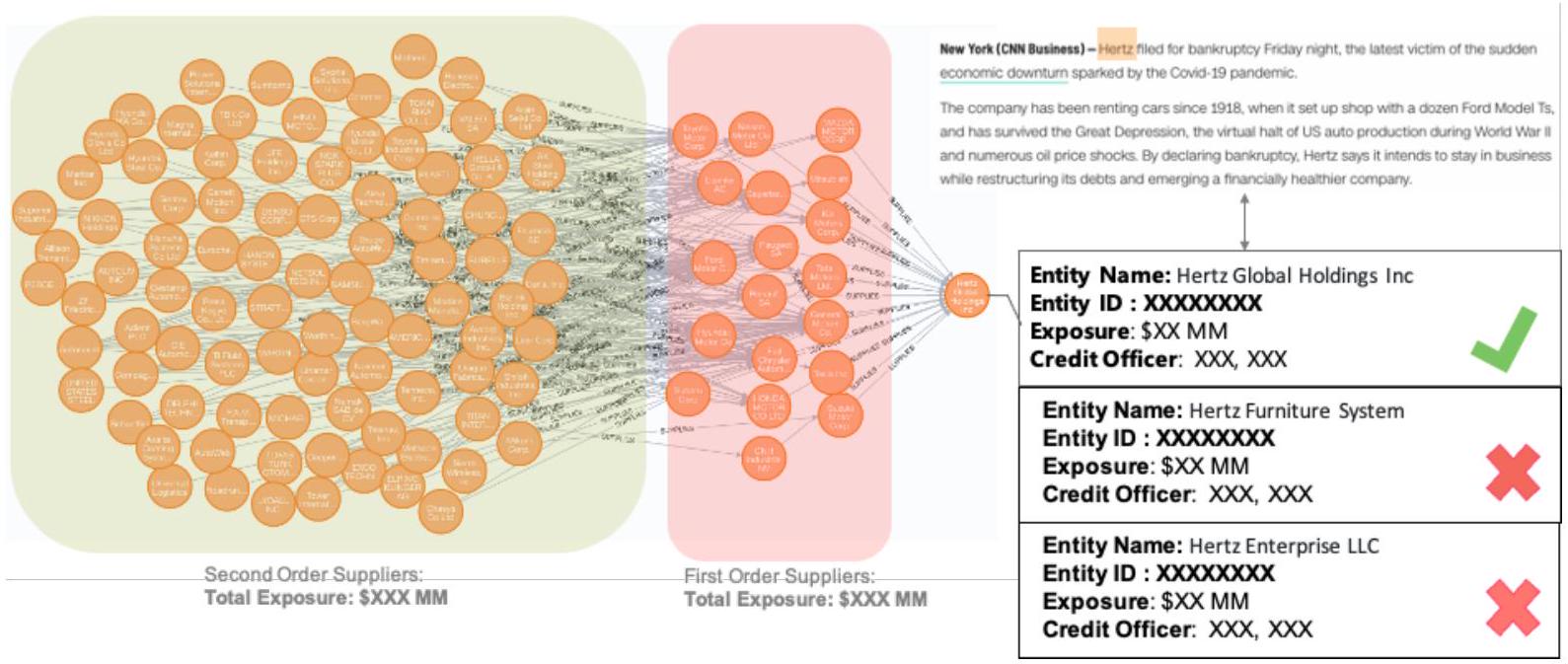}
\captionsetup{labelformat=empty}
\caption{Figure 1: Demonstration of Leveraging Knowledge Graph for News Analytics}
\end{center}
\end{figure}

"Big Apple". Similarly, An entity may have ambiguous meanings. For example, the entity "Sun" can refer to the star at the center of the Solar System or the multinational company, "Sun Microsystem". An entity linking system should be able to disambiguate the entity mentioned in the textual context and identify the mapping entity for each mention.

\section*{2 PROBLEM DEFINITION AND METHODOLOGY}
\subsection*{2.1 Problem Definition}
Given a knowledge graph containing a set of entities $E$ and a collection of text (news) in which a set of named entity mentions $M$ are identified in advance, the goal of entity linking is to map each textual entity mention $m \in M$ to its corresponding entity $e \in E$ in the knowledge base.

Now, we present an example of an entity linking task shown in Figure 2. For the text on the left of the figure, an entity linking system should leverage the available information, such as the context of the name entity mention and the entity information from the knowledge base, to link the named entity mention "Michael Jordan" with the Berkeley professor Michael I. Jordan, rather than other entities, such as the NBA player Michael J. Jordan, the English football goalkeeper Michael W. Jordan, etc..

\subsection*{2.2 Methodology}
In order to accomplish an end-to-end entity linking pipeline, we process our data as Figure 3.

Assume we get a piece of news "... make a decision about fundamental aspects. Joe Adam explains why the government needs...", we apply a NER (Named Entity Resolution) over this news, and detect "Joe Adam" in the text is a person's name, which we want to link to our knowledge graph. Entity linking is inherent of quadratic complexity. Matching every mention across all entities in a knowledge graph is impractical. Here we apply a blocking layer over the entities to reduce the number of performed comparisons. We

\begin{figure}[h]
\begin{center}
  \includegraphics[width=\textwidth]{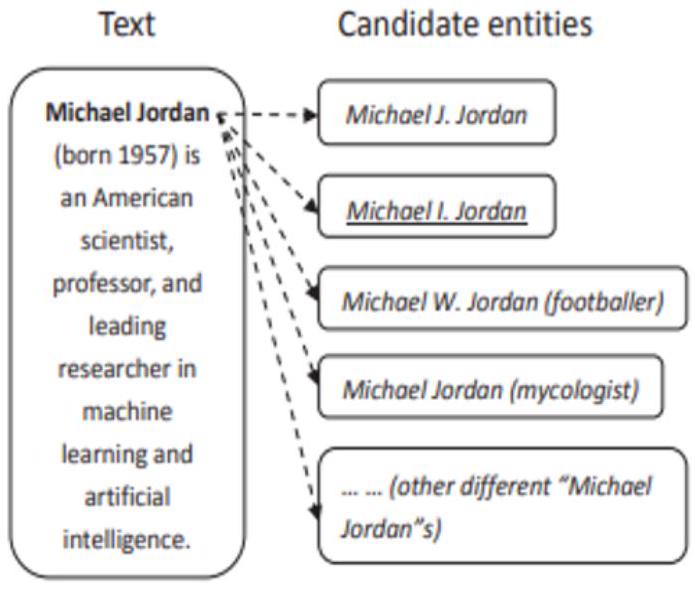}
\captionsetup{labelformat=empty}
\caption{Figure 2: An illustration for the entity linking task}
\end{center}
\end{figure}

leverage a simple but fast fuzzy match in the blocking layer to filter out those impossible matches. In the case of "Joe Adam", entities with names such as "Elon Musk", "Jeff Bezos", "Bill Gates" will be filter out, and entities like "Jo Adam","Joe Ada", "Joseph Adam" etc. will be kept as candidates since they are similar enough. In our settings, we average the similarity score from Cosine Similarity, Levenshtein Similarity, and Jaro Similarity between the mention and entity names, and only keep the entities with at least $50 \%$ similarity as candidates.

After collecting the candidates, we will apply a much more sophisticated entity linking model to decide which entity we will link to the mentioned in the news. Usually, two parts of information determine whether a mention and an entity will be linked or not: Surface Information and Semantic Information. Surface Information describes whether two name strings look alike. For example, "Joe Adam" and "John Adam" should not be linked since "Joe" and

\begin{figure}[h]
\begin{center}
  \includegraphics[width=\textwidth]{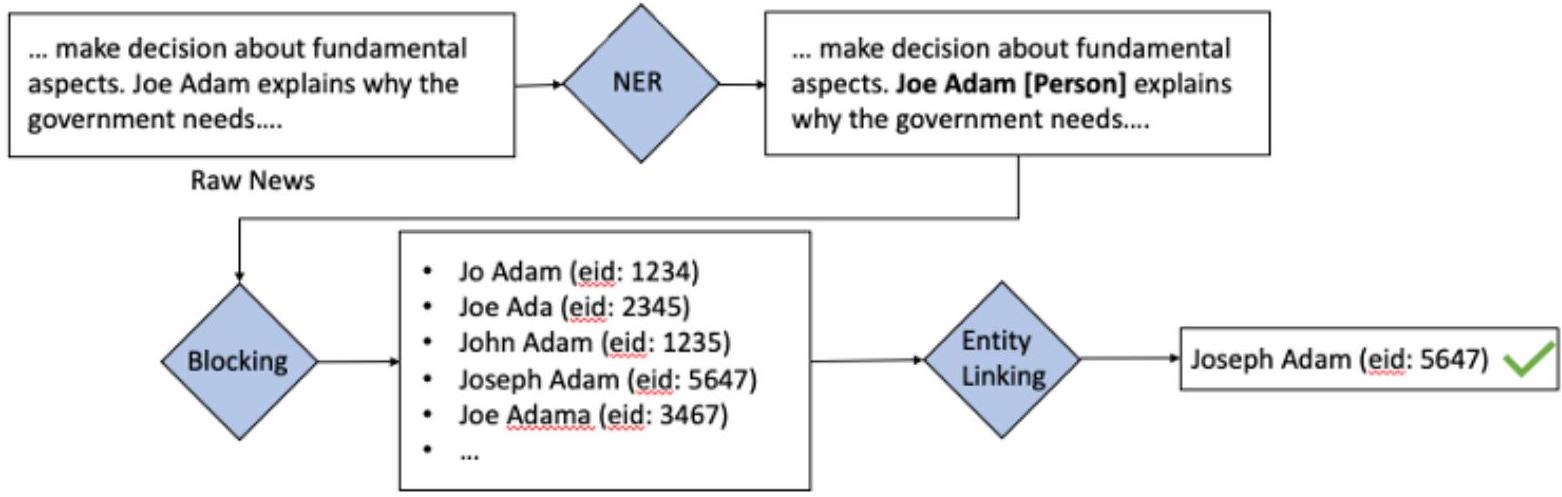}
\captionsetup{labelformat=empty}
\caption{Figure 3: Data Process Pipeline}
\end{center}
\end{figure}

"John" are two different names. Semantic information describes the semantic meaning behind the names. For example, it is hard to tell whether "Joe Adam" and "Joseph Adam" are the same person without collecting more background data, such as birth year, biography, etc. that describes these two names.

To combine these two factors, we've designed a model in Figure 4 to map a mention from a piece of news to an entity in our DaVinci knowledge graph.

Our entity linking model consists of two parts: Surface Match and Semantic Match.

We apply three levels of embedding to capture names' surface patterns. First, we observe that character-level matches are often important: for example, "Joe Adam"= "Joseph Adam" and "Dave Smith" ="David Smith". This pattern requires character-level information to be modeled. Thus, we apply a Char Embedding layer to capture this level of information. Second, word-level information is also important: for instance, "John Adam" != "Joe Adam", "Mary Miller" != "Mark Miller", but "Bill Gates" = "William Gates". While character-level models can capture some of these, it can be difficult for long names with many tokens. Thus, an explicit word-level model is useful. We concatenate the embeddings from the characterlevel model and transform them into word embeddings to capture the information from the word level. At the third level, we ensemble the word embeddings into one embedding representing the entity we are trying to learn: entity-level embedding.

We utilize BERT [1] to extract the semantic information from the news that contains the mention. BERT is a pre-trained natural language processing model released by Google. With a BERT, we can transform a mention string into a computable numerical vector with context information involved. In Figure 4, we can see that the context around "Joe Adam" mentioning about "fundamental aspects", and "government". A BERT model can represent this "Joe Adam" with a numerical vector whose semantic meaning is closer to a government officer.

As for the Entity Embedding described in Figure 4, we pre-trained a TripletNet(shown as Figure 5) model to embed each entity in our\\
knowledge graph. Since the description sentences for one entity are usually short, with which BERT could not work the best, we have designed the Triplet Network shown as Figure 4 to learn embeddings for each of the entities in our knowledge graph. Take "Jo Adam" in Figure 4 as an example, we can find that "Finance" is one of the most representative words from his/her biography. Furthermore, we can find this person has no relationship with the word "Military". Then we train the model in 5) to ensure the Anchor Embedding for "Jo Adam" is closer to Finance's word embedding and further away from Military's word embedding. We leverage fasttext as the pre-trained word embedding model.

After we get both the mention embedding from its context and entity embedding from its biography information, we transform each of them with one layer of linear regression to make them compatible since they come from different embedding methods. Then we concatenate the surface embedding with semantic embedding as one embedding to represent the mention or the entity.

Following the steps described above, we've finished the embedding part. In the next part, we will apply a comparison module to score the possibility these two are linked or not. The comparison module is shown as the gray part in Figure 5. In our current setting, we just adopt a simple two-layer linear regression as the comparison module. We concatenate the mention embedding and entity embedding into one vector, send it to the two-layer linear regression module, and get one score from it. If the final score larger than 0.5 , we will say the mention and the entity is linked, otherwise not linked.

\section*{3 DATA,RESULT AND DISCUSSION}
Our testing entities come DaVinci People Graph ${ }^{2}$. DaVinci is a Knowledge Graph project of the AWM Private Bank. It combines datasets from the internal system and third-party data from vendors such as WealthX, FactSet, into a connected knowledge graph. Our

\footnotetext{${ }^{2}$ \href{https://confluence.prod.aws.jpmchase.net/confluence/display/AWMAIENGINEERING/DaVinci++PB+Knowledge+Graph}{https://confluence.prod.aws.jpmchase.net/confluence/display/AWMAIENGINEERING/DaVinci++PB+Knowledge+Graph}
}\begin{figure}[h]
\begin{center}
  \includegraphics[width=\textwidth]{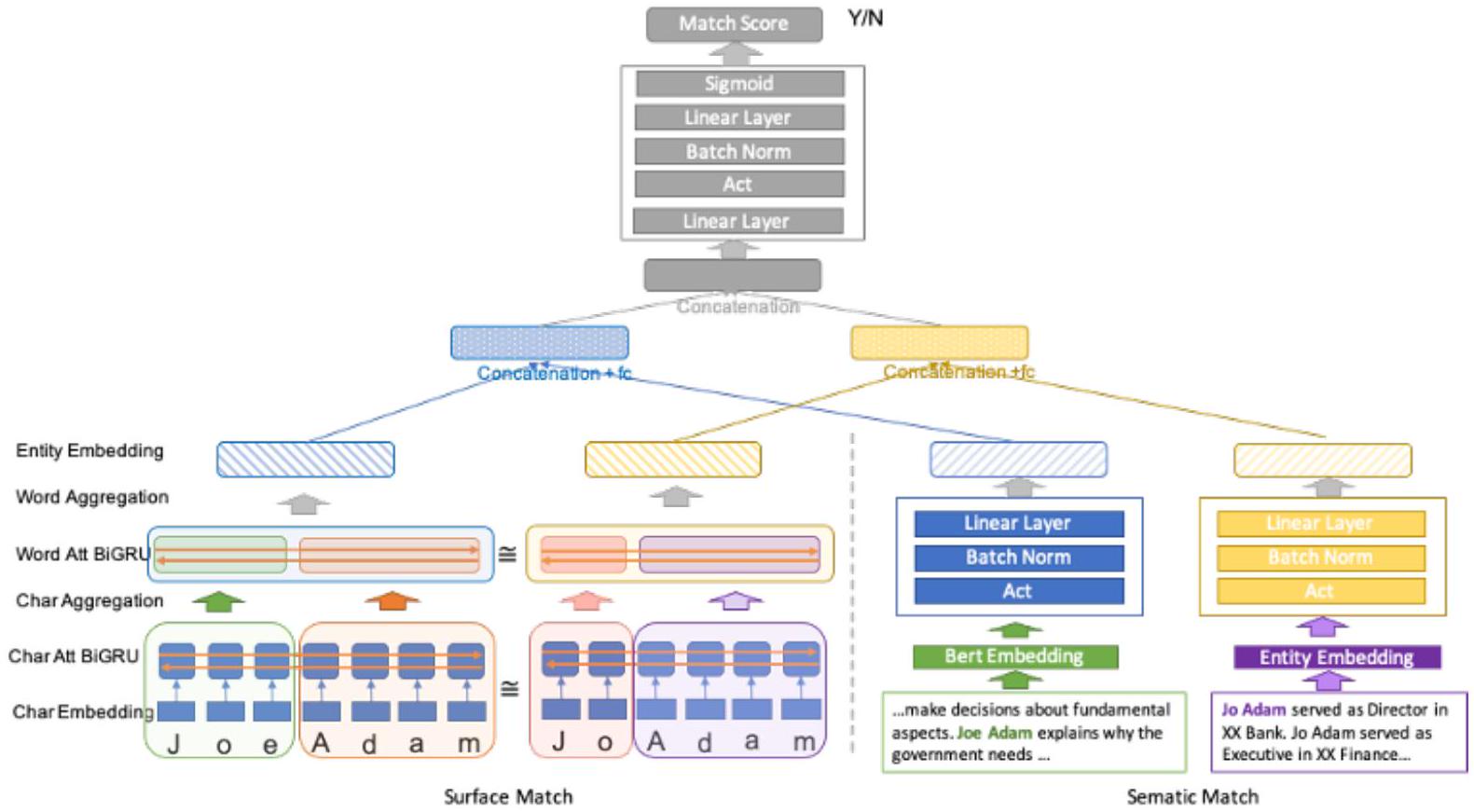}
\captionsetup{labelformat=empty}
\caption{Figure 4: Model Description}
\end{center}
\end{figure}

\begin{figure}[h]
\begin{center}
  \includegraphics[width=\textwidth]{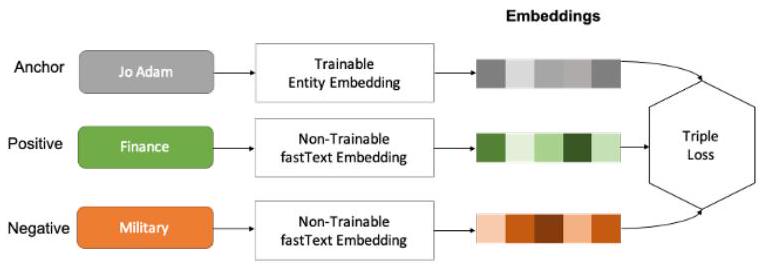}
\captionsetup{labelformat=empty}
\caption{Figure 5: Model Description}
\end{center}
\end{figure}

testing news comes from the Dow Jones news stream. In the following part of this section, we will evaluate our model's performance on DaVinci people graph and Dow Jones news.

\subsection*{3.1 Quantitative Evaluation}
We collected two sets of data from DaVinci and Dow Jones using a fuzzy match score together with a pre-defined threshold. The high confidence data contains highly confident labeled data. Since both people's and company names are mentioned in the article, the mappings between mentions and entities are mostly correct. While the low confidence dataset contains some noise since only people's names are mentioned in the article, there might be some mismatch.

Another one is a noisy dataset, which contains some noisy labels.

\begin{itemize}
  \item High confidence dataset (fuzzy match similarity >= 0.9) people and companies are mentioned in the same article.
  \item Low confidence dataset (fuzzy match similarity >= 0.8) only people names are mentioned, but company names are missing
\end{itemize}

Facebook released its state-of-the-art model, BLINK[6], in 2020. We took the BLINK model as a benchmark over our high confidence dataset, and compared our in-house model with it. Table 1 provides an overview of the two models' performances. In summary, our in-house model performs better than BLINK: we have increased the F1 score more than by 15\%, since our model considers surface and semantic similarity between entities and mentions, while BLINK only considers semantics.

To test whether our in-house model can deal with noisy data, we also test our model's performance over the low-confidence dataset. The performance is shown in Table 2. According to Table 2, our entity linking model performed pretty well on both datasets, even though the low-confidence data has much noise. In addition, we also find that our model has an impressive capability to distinguish mentions and entities that share the same name but are different entities. For example, even a sophisticated fuzzy match will mismatch David Davis (British politician) with David Davis (US businessman), but our entity linking model can distinguish them well.

\subsection*{3.2 Ablation Study}
Our current system, Galileo ${ }^{3}$, is running a fuzzy match for the entity linking problem. However, we observe several issues with fuzzy match results, especially for those people who share similar names but are different entities in the real world. However, we observe that this problem can be treated well with a machine learning entity linking model. Here are some examples we find from our data.

\footnotetext{${ }^{3}$ \href{https://confluence.prod.aws.jpmchase.net/confluence/display/AWMAIENGINEERING/Galileo++News+Analytics+Platform}{https://confluence.prod.aws.jpmchase.net/confluence/display/AWMAIENGINEERING/Galileo++News+Analytics+Platform}
}\begin{table}[h]
\begin{center}
\begin{tabular}{|l|l|l|l|l|l|}
\hline
 & Training/Testing Samples & Accuracy & Precision & Recall & F1 \\
\hline
Facebook BLINK & $1567 / 424$ & 0.8463 & 0.8163 & 0.7187 & 0.7632 \\
\hline
In-House Model & $1567 / 424$ & 0.9327 & 0.9093 & 0.9458 & 0.9272 \\
\hline
\end{tabular}
\captionsetup{labelformat=empty}
\caption{Table 1: Comparison between Facebook's BLINK and Our Model}
\end{center}
\end{table}

\begin{table}[h]
\begin{center}
\begin{tabular}{|l|l|l|l|l|l|l|}
\hline
 & Training/Testing Samples & Accuracy & Precision & Recall & F1 & AUC \\
\hline
Clean Dataset & $5655 / 936$ & 0.9327 & 0.9093 & 0.9458 & 0.9272 & 0.9788 \\
\hline
Noisy Dataset & $7622 / 498$ & 0.8808 & 0.8854 & 0.8543 & 0.8696 & 0.9571 \\
\hline
\end{tabular}
\captionsetup{labelformat=empty}
\caption{Table 2: In-House Model's performance over different datasets}
\end{center}
\end{table}

\section*{Example 1:}
\begin{itemize}
  \item Fuzzy Match: Match (sim\_score: 1.0)
  \item EL Model: Not Match
  \item Mention: David Davis
  \item Entity: David Davis
  \item Context: ... David Davis Pictures show the burning wrecks of Russian aircraft and the terrified faces of Putin's men as they were detained after ejecting from their planes ...
  \item Entity Description: ... David Hammeken Davis,also known by his nickname as Dave Davis.David Hammeken Davis primarily works for Sweetwater Farm...\\
In this example, David Davis mentioned in the news is a British politician, while the matched David Davis in the knowledge graph is a US businessman who owns Sweetwater Farm. They share the same name but they are two different persons. Fuzzy match treat them as the same person, but our model can distinguish them since they have different semantic contexts.
\end{itemize}

\section*{Example 2:}
\begin{itemize}
  \item Fuzzy Match: Not Match
  \item EL Model: Match
  \item Mention: Christopher Nolan
  \item Entity: Christopher Nolan
  \item Context: ... Christopher Nolan "I don't think they are making them an elevated art form" Cronenberg said of prominent directors taking on superhero movies ." I think it ś still Batman running around in a stupid cape, ...
  \item Entity Description: ... Christopher Jonathan James Nolan,also known by his nickname as Chris Nolan Christopher Edwards Nolan Christopher Johnathan James Nolan. Christopher Jonathan James Nolan primarily works for Syncopy......\\
The two Christoper Nolan are the same person, who is the famous film director. However, since the mention-context does not hint at the company name, the fuzzy match missed this. However, our entity linking model finds this match.
\end{itemize}

\section*{4 RELATED WORK}
Entity Linking is a traditional NLP task and has been discussed for years by different perspectives. Recently, deep learning methods [2,4,7] have achieved wide success in this task. Our model is also a deep learning model. Facebook's BLINK[6] has achieved state-of-the-art performance,but BLINK does not consider surface\\[0pt]
features, failing to map some mention-entity pairs with surface differences. The model described in the study [5] has achieved promising performance as an end-to-end model. According to this study[5], authors apply two levels of Bi-LSTM to embed characters into words and words(character embedding+pre-trained word embedding) into mentions. With calculating the similarity between a mention vector and a pre-trained entity vector (this study adopted the pre-trained entity embedding from study[3] directly) to decide whether they are matched or not. Similarly, this study does not consider surface features. In addition, Bi-LSTM is hard to train, especially we don't have much labeled training data.

\section*{5 CONCLUSION AND FUTURE WORK}
In this paper, we created and deployed a novel neural entity linking model to leverage our knowledge graph DaVinci in news analysis. Our entity linking model utilizes a hierarchical surface embedding to capture character/word level information, and uses deep learning model to capture semantic information. With an extensive experiments and careful manual investigation, we have shown the superiority of this method. We are currently in process of deploying this model in our Galileo system.

In the next step, we will spare effort to deal with nickname issues to improve model performance. Later, we will put this model into production to serve the Galileo News Analytics system.

\section*{REFERENCES}
[1] Jacob Devlin, Ming-Wei Chang, Kenton Lee, and Kristina Toutanova. 2018. Bert: Pre-training of deep bidirectional transformers for language understanding. arXiv preprint arXiv:1810.04805 (2018).\\[0pt]
[2] Matthew Francis-Landau, Greg Durrett, and Dan Klein. 2016. Capturing semantic similarity for entity linking with convolutional neural networks. arXiv preprint arXiv:1604.00734 (2016).\\[0pt]
[3] Octavian-Eugen Ganea and Thomas Hofmann. 2017. Deep joint entity disambiguation with local neural attention. arXiv preprint arXiv:1704.04920 (2017).\\[0pt]
[4] Hongzhao Huang, Larry Heck, and Heng Ji. 2015. Leveraging deep neural networks and knowledge graphs for entity disambiguation. arXiv preprint arXiv:1504.07678 (2015).\\[0pt]
[5] Nikolaos Kolitsas, Octavian-Eugen Ganea, and Thomas Hofmann. 2018. End-toend neural entity linking. arXiv preprint arXiv:1808.07699 (2018).\\[0pt]
[6] Martin Josifoski Sebastian Riedel Luke Zettlemoyer Ledell Wu, Fabio Petroni. 2020. Zero-shot Entity Linking with Dense Entity Retrieval. In EMNLP.\\[0pt]
[7] Yaming Sun, Lei Lin, Duyu Tang, Nan Yang, Zhenzhou Ji, and Xiaolong Wang. 2015. Modeling mention, context and entity with neural networks for entity disambiguation. In Twenty-fourth international joint conference on artificial intelligence.

\end{document}